\documentclass[runningheads,a4paper]{llncs}
\usepackage{amssymb}
\usepackage{graphicx}
\usepackage{scrextend} 
\usepackage{url}

\usepackage{cite} 
\usepackage{amsmath} 

\usepackage{algorithm}
\usepackage[noend]{algpseudocode}

\urldef{\mailsa}\path|p.ambrosini@erasmusmc.nl|
\newcommand{\keywords}[1]{\par\addvspace\baselineskip
\noindent\keywordname\enspace\ignorespaces#1}

\usepackage{fancyhdr}
\begin{document}
	\mainmatter
	\title{Fully Automatic and Real-Time Catheter Segmentation in X-Ray Fluoroscopy}
	\titlerunning{Catheter Segmentation in X-ray}
	\author{Pierre Ambrosini\inst{1,3} \and Daniel Ruijters\inst{2} \and Wiro J. Niessen\inst{1,3,4} \and Adriaan Moelker\inst{3} \and Theo van Walsum\inst{1,3}}
	\authorrunning{Ambrosini et al.}
	
	\institute{Biomedical Imaging Group Rotterdam, Department of Medical Informatics,\\Erasmus MC, Rotterdam, The Netherlands\\
	\mailsa
	\and Philips Healthcare, Interventional X-ray Innovation, Best, The Netherlands
	\and Department of Radiology and Nuclear Medecine,\\Erasmus MC, Rotterdam, The Netherlands
	\and Imaging Physics, Faculty of Applied Sciences,\\Delft University of Technology, Delft, The Netherlands}
	
	\maketitle
	\thispagestyle{fancy}
	\pagestyle{empty}
	
	\setcounter{footnote}{0}
	
	\begin{abstract}
		Augmenting X-ray imaging with 3D roadmap to improve guidance is a common strategy.
		Such approaches benefit from automated analysis of the X-ray images, such as the automatic detection and tracking of instruments.
		In this paper, we propose a real-time method to segment the catheter and guidewire in 2D X-ray fluoroscopic sequences. The method is based on deep convolutional neural networks. The network takes as input the current image and the three previous ones, and segments the catheter and guidewire in the current image. Subsequently, a centerline model of the catheter is constructed from the segmented image.
		A small set of annotated data combined with data augmentation is used to train the network.
		We trained the method on images from 182 X-ray sequences from 23 different interventions. On a testing set with images of 55 X-ray sequences from 5 other interventions, a median centerline distance error of 0.2 mm and a median tip distance error of 0.9 mm was obtained.
		The segmentation of the instruments in 2D X-ray sequences is performed in a real-time fully-automatic manner.

	\end{abstract}

	\section{Introduction}
		Minimally invasive procedures are generally preferred over open surgery interventions, as these localized and accurate interventions lead to less trauma and shorter recovery times than conventional procedures. Minimally invasive procedures require real-time imaging to visualize the relevant anatomy and the instruments. Particularly, in catheterization procedures, a catheter \footnote{For clarity, in the remainder of the paper, the word ``catheter'' also refers to the micro-catheter and guidewire instruments. Although they have quite different appearances, they are handled altogether as one instrument in the proposed method.} is inserted into the body via the vasculature and fluoroscopic imaging is used to continuously visualize the catheter.
		The vasculature is only visible in X-ray images when contrast agent is injected, and contrast agent is used sparingly because of its toxic nature. Therefore, recent approaches for virtual roadmapping that permit the visualization of a 3D vessel tree from pre-operative images have been presented \cite{ambrosini20153d,ruijters2011validation}.
		Such methods benefit from automated extraction of the instruments from fluoroscopic images.
		The purpose of this work was therefore to develop and evaluate a method that segments fully automatically the catheter in 2D single-plane X-ray fluoroscopic sequences in real-time.
		
		Automatic catheter segmentation is not straightforward, as the catheter is a thin, moving structure with low contrast in noisy images.
		Segmentation methods for electrophysiology (EP) electrodes and EP catheter in 2D X-ray images have been reported \cite{baur_cathnets_2016,wu2015fast}. EP electrodes are clearly visible, and their location is often used to obtain a full segmentation of EP catheters.
		Segmentation of catheters without features such as electrodes has been studied less frequently. Most methods enhance the instruments with Hessian-based filters, which are followed by a spline fitting approach, starting from the catheter shape of the previous frame \cite{baert2003guide,slabaugh2007variational,wang_robust_2009,hauke2012interventional,chang2016robust,chen2016guidewire}.
		These methods have two drawbacks: the first frame of the fluoroscopic sequence has to be manually annotated and the curvature and length of the catheter should not change much between frames.
		\cite{hauke2012interventional,chang2016robust} propose semi-automatic methods to segment the first frame.
		Recently, a fully automatic method using directional noise reduction and path extraction, with segments and similarity from the previous frame cost function, has been proposed \cite{wagner2016guidewire}.
		The method was evaluated on the last frame of 7 sequences from one canine study on which it performs well; it is, however, not a real-time method.
		We summarize in Table~\ref{table:tableRelatedWork} the methods proposed in the literature in order to show results they obtained. Note that the results, the accuracy metrics and computation times cannot be directly compared but they give an idea of the performances.

		\begin{table}
			\caption{Summary of the methods in the literature and the method of this paper.}
			\begin{minipage}{\textwidth}
				\centering
				\setlength\tabcolsep{3pt} 
				\begin{tabular}{ccccccc}
					References & \begin{tabular}[x]{@{}c@{}}Fully\\Auto.\end{tabular} & Time & Accuracy & Tip Accuracy\\
					\hline
					2003 Baert et al.\cite{baert2003guide} & No & 5 s & mean 0.9 px\footnote{\label{noteSuccess}The failed segmentations are not included in the evaluation} & mean $< 2$ mm\footref{noteSuccess} \\
					2007 Slabaugh et al.\cite{slabaugh2007variational} & No & 175 ms & - & - \\
					2009 Wang et al.\cite{wang_robust_2009} & No & 500 ms & mean 2 px (0.4 mm) & mean 5.4 px \\
					2012 Heibel et al.\cite{hauke2012interventional} & No & 60 ms & mean 0.8-3.9 px & - \\
					2016 Chang et al.\cite{chang2016robust} & No & - & - & - \\
					2016 Chen et al.\cite{chen2016guidewire} & No & - & mean 2.1 px (0.5 mm) & - \\
					2016 Wagner et al.\cite{wagner2016guidewire} & Yes & $> 1$ min & mean 0.5 mm & - \\
					This work & Yes & 125 ms & median 0.2 mm & median 0.9 mm \\
				\end{tabular}
			\end{minipage}
			\label{table:tableRelatedWork}
		\end{table}
	
		Our method utilizes deep convolutional neural networks (CNNs) for the segmentation. CNNs have been demonstrated to be very effective in image classification and image segmentation \cite{long_fully_2015}, also in case of medical images with a limited set of annotations \cite{milletari2016v,ronneberger_unet_2015}.
		Ronnerberger et al. \cite{ronneberger_unet_2015} introduced an end-to-end biomedical imaging segmentation network called U-net: a model with a fully convolutional part (downsampling part) and a deconvolutional part (upsampling part) which outputs after appropriate thresholding a binary segmented image.
		Extensive data augmentation enables the neural network to generalize well, even in case of small training sets.
		Several improvements w.r.t. the network and the training process have been introduced more recently.
		The downsampling (resp. upsampling) part has been shown to be more effective with strided convolution (resp. transposed convolution) than with max pooling \cite{milletari2016v,springenberg2014striving}. Strided convolution enables to learn how the features should be downsampled/upsampled.
		Moreover, batch normalization \cite{ioffe_batch_2015} and residual learning \cite{he_deep_2015} have been proposed to improve training convergence.

		2D X-ray fluoroscopic images are very noisy and are used in liver catheterization procedures to guide catheter inside the liver vessel tree. The catheter used does not have specific features, such as electrodes, nor specific shapes that may facilitate their segmentation. The framerate is around 7Hz, so, during the catheter manoeuvre, its shape and length may change considerably between two consecutive frames. Tracking the catheter over time is therefore quite challenging.
		In this paper, we propose a fully automatic segmentation method based on the U-net model combined with recent strategies to improve the training of the network, such as batch normalization, residual learning and a data augmentation scheme to increase the size of the training dataset. The catheter centerline is then extracted using skeletonization and linking of the extracted branches.
		Our work differs from previous approaches in that we introduce a fully automatic approach that can be run in real-time.

	\section{Method}
		The catheter is segmented using the CNN and then the centerline is extracted from the result of the CNN using skeletonization and subsequent linking of branches.
	
		\subsection{Data}
			A 2D X-ray image sequence is a set of $s$ 2D images $S = \{I_1, I_2, \ldots, I_{s}\}$. Each image $I_i$ is associated with a binary image $B_i$ where the catheter pixels have a value of 1 and the background pixels 0. We also associate the output from the neural network, the image prediction $B^{p}_i$, where each pixel is between 0 and 1, and a pixel closer to 1 is considered as a catheter pixel.
			Our neural network model is trained to predict a mask $B^{p}_i$, given an image $I_i$ as input.

		\subsection{CNN Model}
\begin{figure*}
				\begin{minipage}[c]{.30\textwidth}
					\caption{Neural network model (top): $I_i, I_{i-1}, \ldots I_{i-3}$ images are fed into the model and segmented image $B^{p}_i$ is predicted. The model is composed of n-conv block at each layer. An n-conv block is $n$ consecutive convolutions of input features (with $f$ filters, height $h$ and witdh $w$) with a residual connection to the output \cite{milletari2016v,he_deep_2015} (bottom).}
					\label{fig:cnn_ipcai}
				\end{minipage}
				\begin{minipage}[c]{.70\textwidth}
					\centering
					\includegraphics[width=0.80\textwidth]{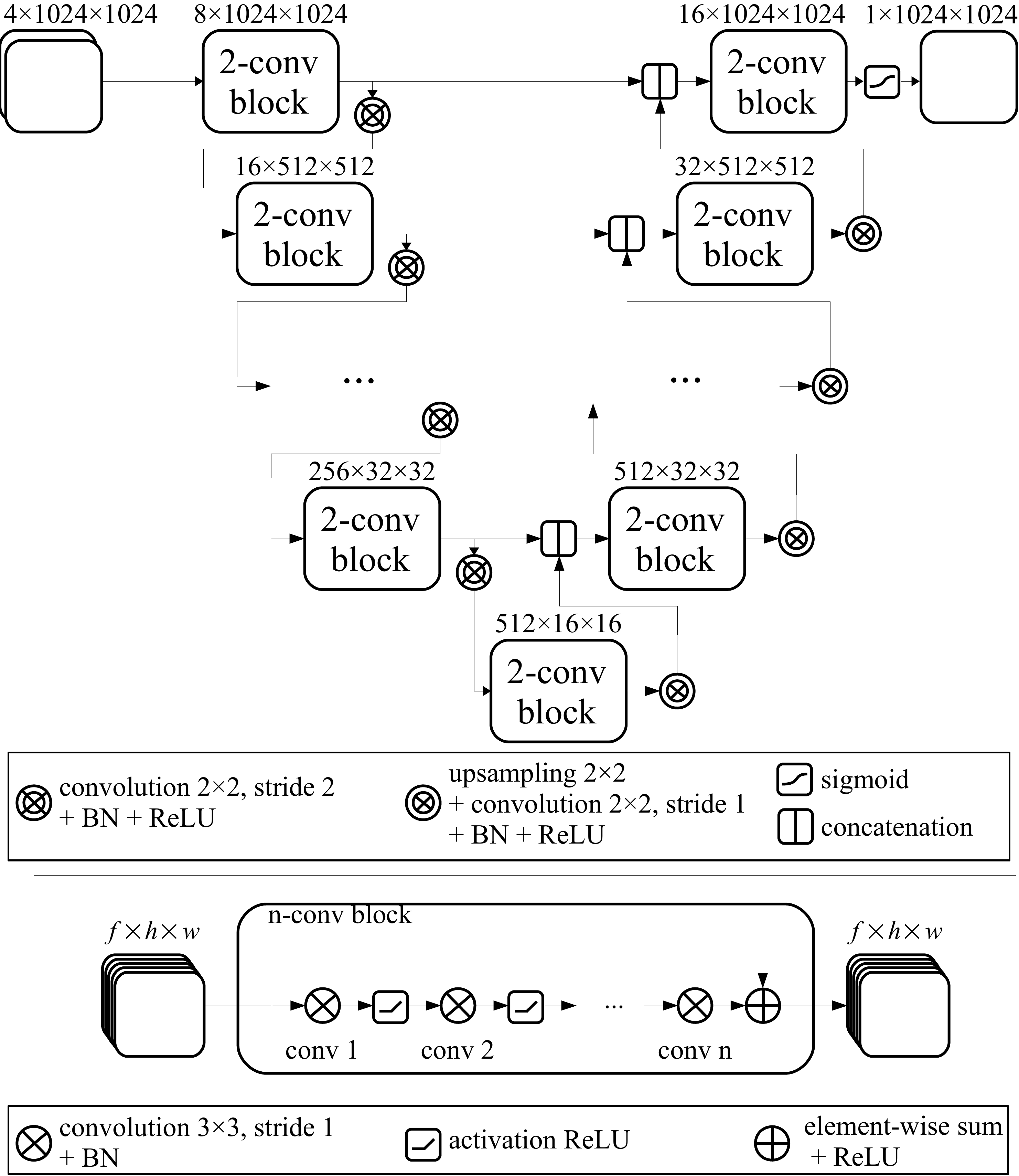}
				\end{minipage}
			\end{figure*}

			The model is an adapted version of the well-known U-net model \cite{ronneberger_unet_2015}.
			The input of the model is the current image $I_i$ and previous frames $I_{i-1}, \ldots I_{i-3}$. The output is the image prediction $B^{p}_i$.
			For the network topology, see Figure~\ref{fig:cnn_ipcai}.
			To improve convergence speed during training, we add batch normalization (BN) after every convolution \cite{ioffe_batch_2015}. In order to also learn how to downsample/upsample the features we use strided convolutions \cite{milletari2016v,springenberg2014striving}.
			To prevent overfitting we add dropout at the end of the two last blocks in the downsampling part \cite{srivastava2014dropout}.
			
		\subsection{Training}
				The loss function is based on the Dice overlap metric \cite{milletari2016v} between the ground truth mask $B_i$ and the output image $B^{p}_i$ of the model, defined as:
				\begin{equation}L_\mathrm{Dice}(B_i, B^{p}_i) = - \frac{2 \sum\limits_{k} {B_{i}}_k {B^{p}_{i}}_k}{\sum\limits_{k}({B_{i}}_k) + \sum\limits_{k}({B^{p}_{i}}_k)}\end{equation}
				
				\noindent where ${B_{i}}_k$ and ${B^{p}_{i}}_k$ are respectively the pixels of the mask $B_i$ and the output image $B^{p}_i$.

				In order to have more data to train and to generalize well, large data augmentation is used during the training.
				Data augmentation is done on the fly. For every image in the set of training images, there is a $50\%$ probability to augment the image.
				If the image is augmented, we apply all of the following transformations to both the X-ray image inputs and the corresponding binary image output:
				50\% probability on a horizontal and vertical flip, a random rotation (around the image center) in a range of $\pm9$ degrees, a random scale with a factor in a range of 0.9 to 1.1, a random horizontal and vertical translation in a range of $\pm16\%$ of the image size, a random intensity shift with a factor in a range of $\pm0.07$ (in normalized image between 0 and 1) and a Gaussian noise with $\sigma = 0.03$.
				
		\subsection{Centerline extraction}
\begin{figure}
				\centering
				\includegraphics[width=0.99\textwidth]{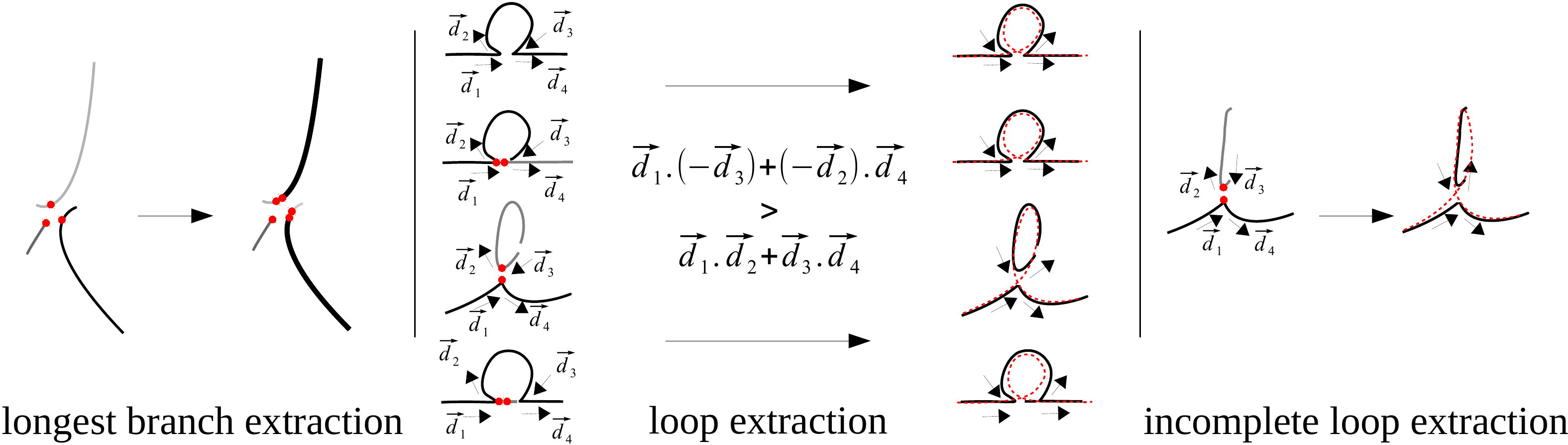}
				\caption{The three steps for the branches linking. Branches are depicted with different colors and connection points are in red. The red dash curve is the centerline after spline smoothing. The first step links the longest branch parts, the second step links and corrects the loops and the last step links the remaining branches longer than $B_{\mathrm{min}}$, considered as an incomplete loop or a straight loop due to foreshortening.}
				\label{fig:centerlineExtraction}
			\end{figure}

			The output of the neural network is first thresholded with a threshold $\alpha$ (between 0 and 1) and then skeletonized \cite{zhang1984fast}.
			Next, the branches (ordered sets of pixels) are determined based on connectivity.
			Connection points are created between close branches. If the closest points between two branches are within a distance $D_{\mathrm{max}}$ pixels, we consider this to be a possible connection, and there can be only one connection between two particular branches.
			Then, to link the branches, three steps are done (Fig.~\ref{fig:centerlineExtraction}). First, for each connection, we divide and merge branches in order to have the longest branches. Second, loops are detected and merged following the direction at the crossing point.
			We have a loop in a branch when two points, within a distance $D_{\mathrm{max}}$ pixels, have their distance along the branch of at least $B_{\mathrm{min}}$ pixels.
			Before the third step, the first and second steps are repeated a second time with a distance $D'_{\mathrm{max}}$ superior to $D_{\mathrm{max}}$.
			Finally, in the last step, the remaining connected branches larger than $B_{\mathrm{min}}$ pixels are considered as incomplete loops or straight loops due to foreshortening. We process them similarly as the second step by closing the two endpoints of their branch.
			When all the potential links have been processed, we keep the longest connected set of branches and choose amongst the two endpoints the farthest from the image border as the tip of the catheter centerline. Finally, the centerline is smoothed by fitting a spline.
			
	\section{Experiments and results}
		
			2D single plane X-ray fluoroscopic sequences have been acquired during 28 liver catheterization procedures in three different hospitals (Erasmus MC, Rotterdam, the Netherlands; H\^opitaux Universitaires Henri Mondor, Cr\'eteil, Paris, France; and Ospedale di Circolo e Fondazione Macchi, Varese, Italy) with angiographic C-arm systems (Xper Allura, Philips Healthcare, Best, the Netherlands).
			Every image is normalized on the range $[0,1]$, mapping the intensities between the 2\textsuperscript{nd} and the 98\textsuperscript{th} percentile of the image histogram.
			In 182 sequences from 23 procedures, we manually segmented the catheter in four consecutive images by annotating points and fitting a spline. From the catheter spline we constructed the binary segmented image using a dilation operator with a $5\times5$ pixel kernel.
			These sequences were used as training data for the parameter optimization.
			In 55 sequences from the other 5 procedures, we also segmented four frames per sequence.
			These 55 sequences will be used as testing data after the model optimization and training.
			
			The loss function $L_{\mathrm{Dice}}$ of the model is optimized using stochastic gradient descent with a learning rate of 0.01, a decay of $5.10^{-4}$ and a momentum of 0.99.
			Following the training, we set the threshold $\alpha$ to 0.01, the maximum distance to connect two branches during the first pass (resp. second pass) $D_{\mathrm{max}}$ to 5 pixels (resp. $D'_{\mathrm{max}}$ to 20 pixels) and the minimum loop length $B_{\mathrm{min}}$ to 30 pixels.
			Using an Nvidia GTX 1080, the average time to segment one image was 125 ms which is suitable for real-time processing.
			Our method is publicly available \footnote{Available at \textit{https://github.com/pambros/CNN-2D-X-Ray-Catheter-Detection}}.

\begin{figure*}
				\begin{minipage}[c]{.355\textwidth}
					\caption{Tip distance error, ground truth to segmented centerline distance error and segmented centerline to ground truth distance error (in mm) on the test set: 4 frames per sequence on 55 sequences.}
					\label{fig:testDiceOneFrame}
				\end{minipage}
				\begin{minipage}[c]{.645\textwidth}
					\centering
					\includegraphics[width=0.59\textwidth]{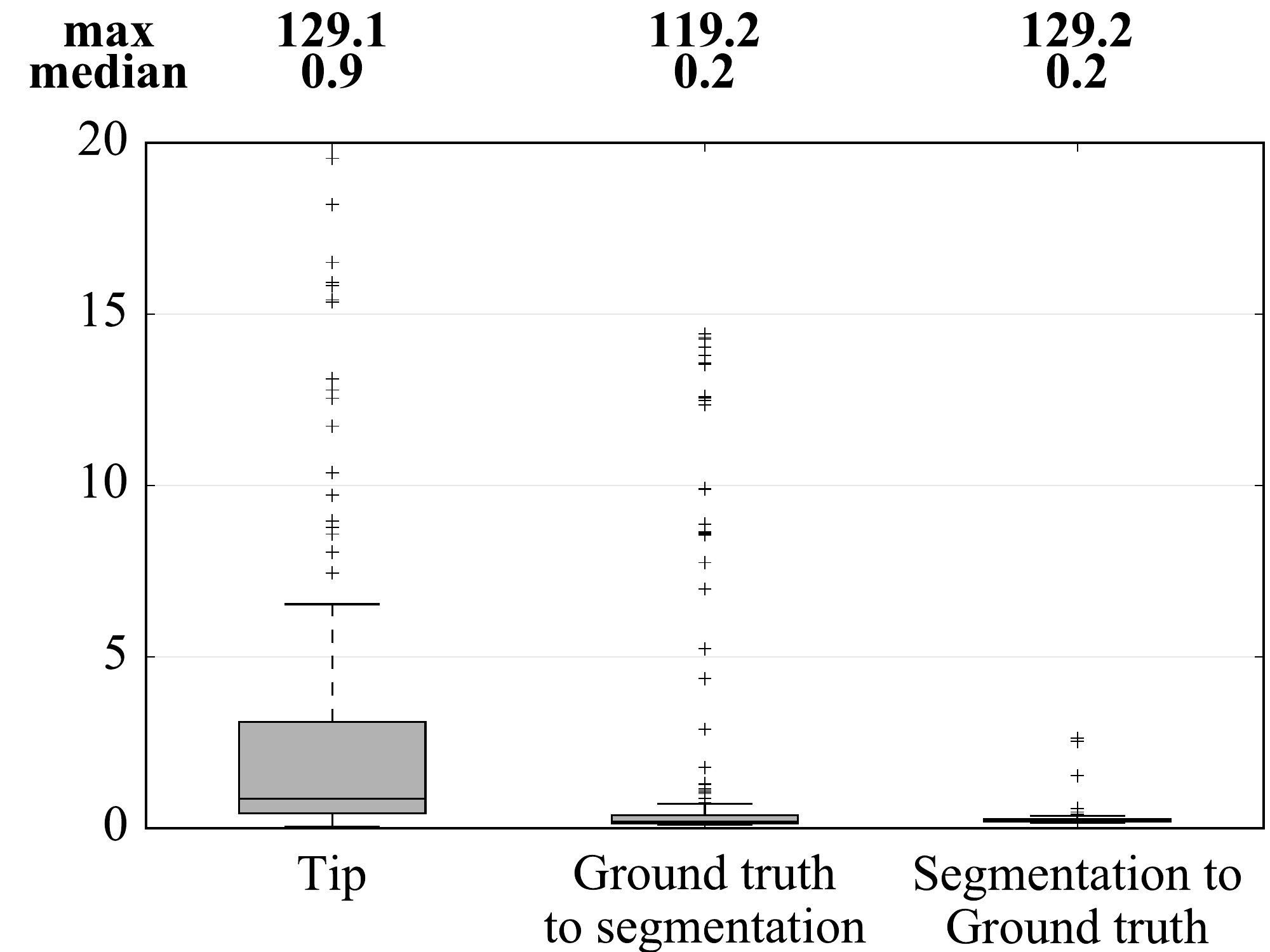}
				\end{minipage}
			\end{figure*}

\begin{figure}
				\centering
				\includegraphics[width=0.19\textwidth]{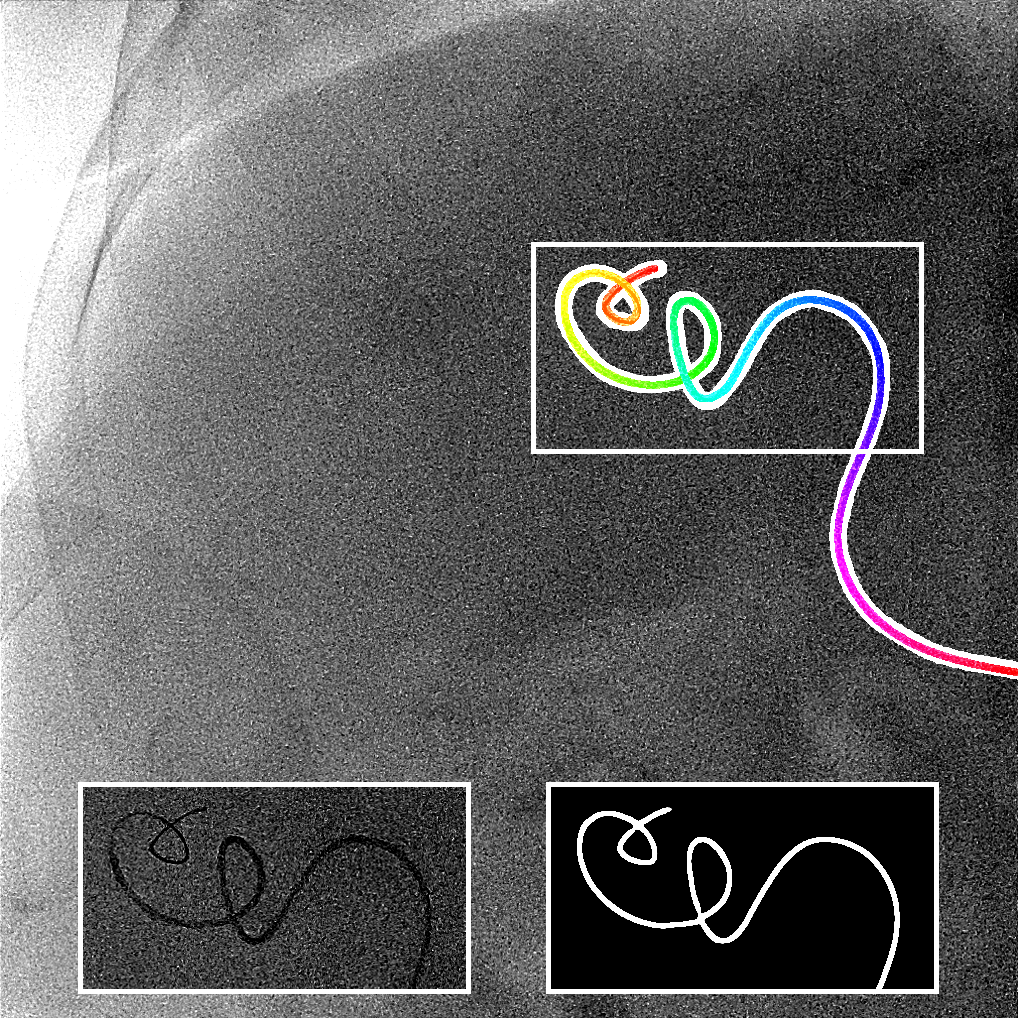}
				\includegraphics[width=0.19\textwidth]{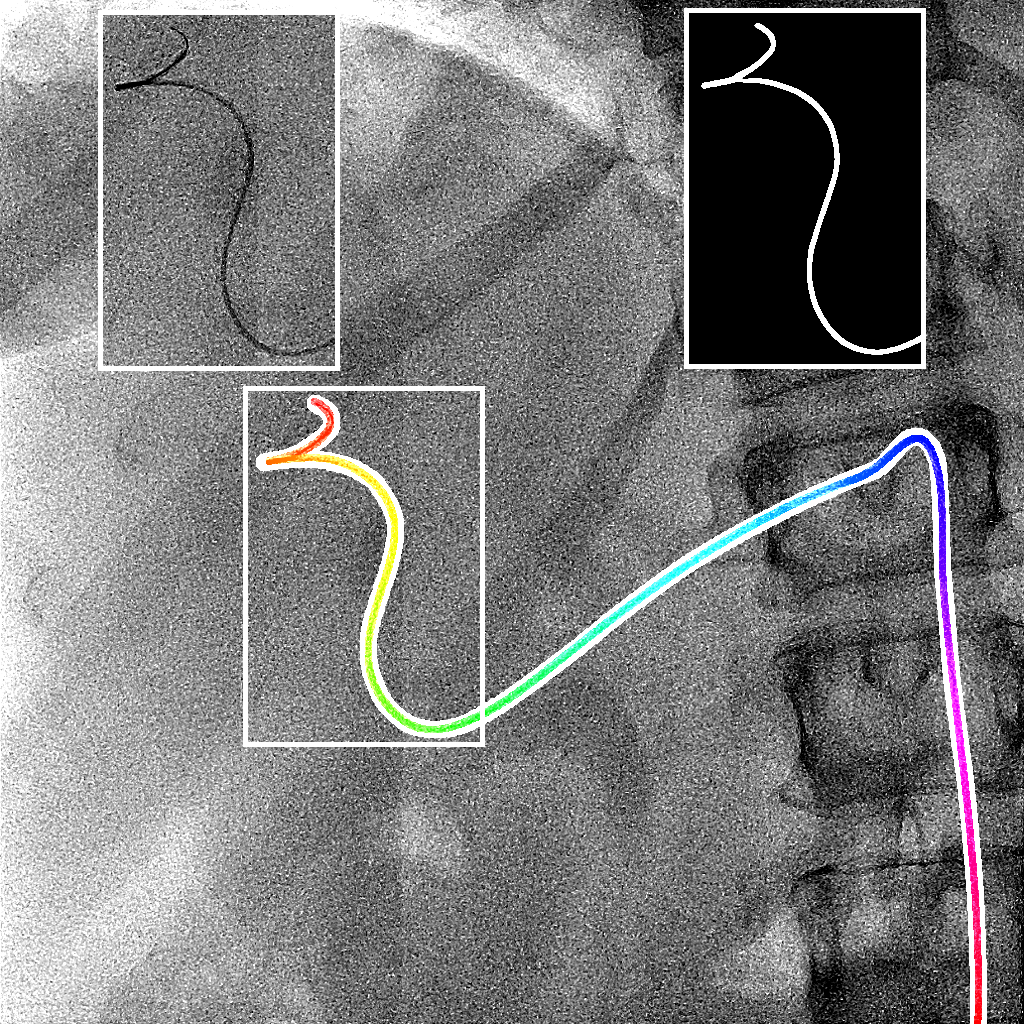}
				\includegraphics[width=0.19\textwidth]{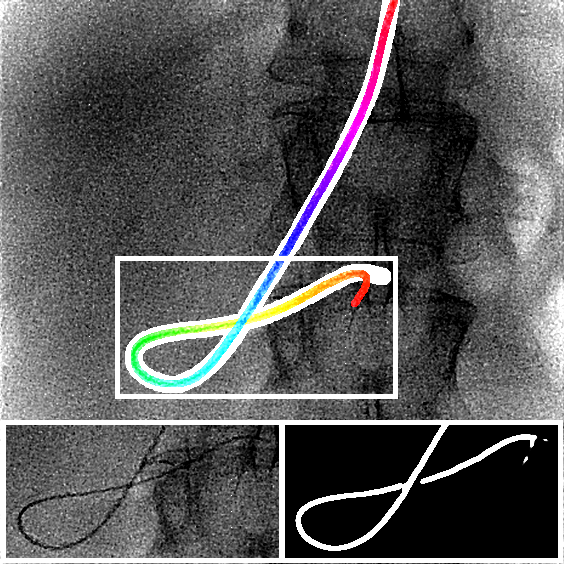}
				\includegraphics[width=0.19\textwidth]{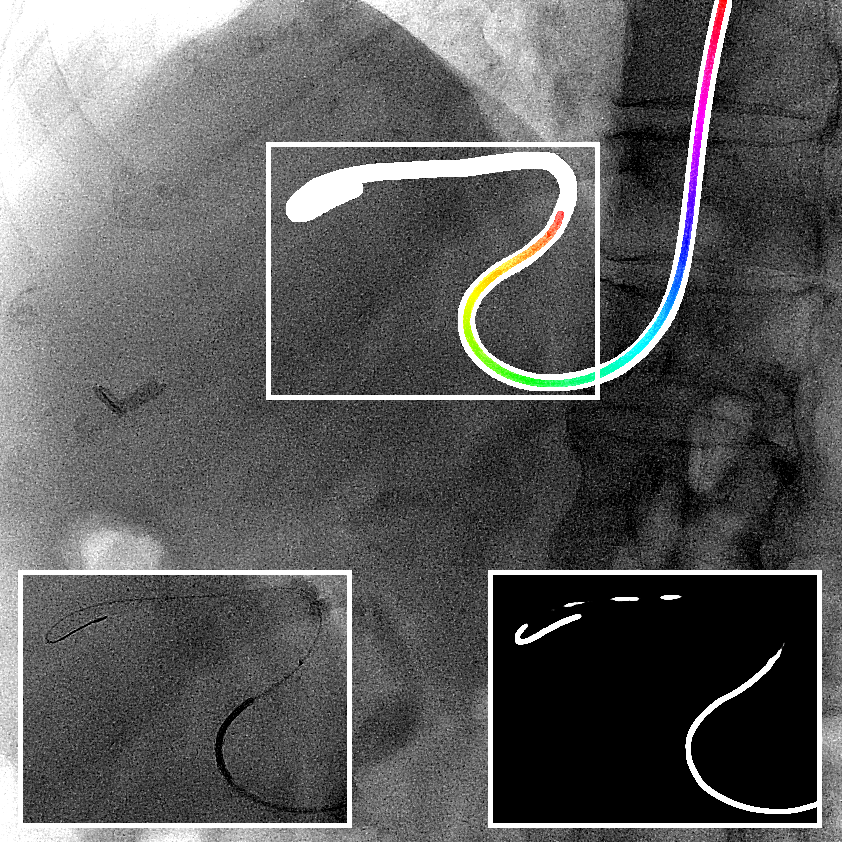}
				\includegraphics[width=0.19\textwidth]{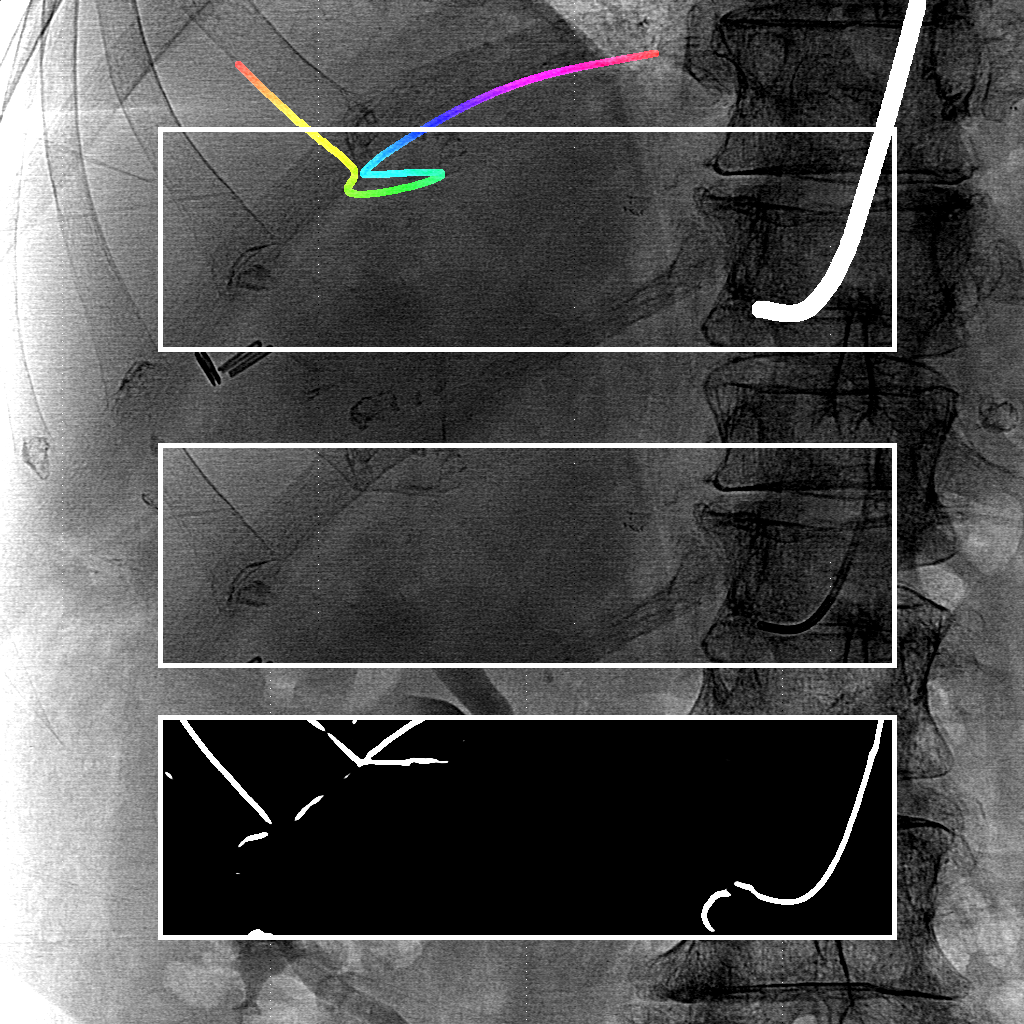}
				\caption{5 segmented frames of the test set. The segmented centerline (dilated for visual purpose) appears from red (tip) to rose. The ground truth (thicker) appears in white. The windows show the original image and the output of the neural network.}
				\label{fig:testExample}
			\end{figure}

			We evaluate using the tip distance error (i.e. the distance between the annotated catheter tip and the tip of the segmented catheter), and the average distance between the manually segmented catheter and the automatically segmented catheter.
			Figure \ref{fig:testDiceOneFrame} shows the tip and catheter distances results.
			We compute the precision of the tip between consecutive frames. The median, average, minimum and maximum of the standard deviation per sequence of the tip distance error are respectively 0.7 mm, 4.9 mm, 0.1 mm and 55.7 mm.
			Five examples of segmentation are shown in Figure~\ref{fig:testExample}. In the third frame, the segmentation is going too far and follows part of the vertebrae. The fourth frame misses the proximal part of the catheter. The last frame is the only sequence with significant false positives.
			It is less noisy because it has been acquired with higher radiation dose.
			The neural network was not trained for such sequence.

	\section{Discussion and conclusion}
	
		We proposed a fully automatic method to segment catheter on 2D X-ray fluoroscopic images using CNNs. The segmentation on testing data gives a median tip distance error of 0.9 mm and a median centerline distance error of 0.2 mm where 85\% of the frames have less than 1 mm of centerline distance error. We note that the distance errors are in mm at the X-ray detector scale. The real distance errors at the patient scale are smaller.

		Very few images have false positives after the CNN segmentation. Therefore, we can use simple criteria to extract the catheter centerline from the CNN segmentation. The results show that it works well and can handle self-intersections.
		The main problem in the extracted catheters are sometimes large gaps in the segmentation due to false negatives. 
		As a consequence, occasionally the proximal part of the catheter is missing.
		With a larger training set, the model is expected to generalize better.

		Previous studies show a higher success rate, probably because they manually initialize the tracking process.
		We, in contrast, do not employ a tracking approach. Whereas it is clear that a stronger incorporation of the time dimension (beyond using consecutive frames in the segmentation) may provide a more robust result, our current results demonstrate that even without tracking good results can be obtained.
		A major advantage of not utilizing tracking is that the method is not hampered by previously incorrectly segmented frames, and thus automatically can recover from previous failures.

		The catheter and guidewire have different thickness and appear quite differently on fluoroscopic images. Whereas we trained one network that segments both, it could be interesting to use two different models and retrospectively combine their results to obtain a more accurate segmentation.
		The current model is using the previous images to segment the catheter but it could also be useful to use the previous segmentation in the model.
		Both strategies are future work.

		To conclude, we developed and evaluated a CNN-based-approach to fully automatically segment catheters in live fluoroscopic images. With execution times within 125 ms, this method is promising for use in real-time catheter detection.
		\vspace*{0.23cm}
	
		\noindent\textbf{Conflict of interest.}
		The authors declare that they have no conflict of interest.

	\bibliography{Library}
	\bibliographystyle{splncs03}
\end{document}